# An Adaptive Alternating-direction-method-based Nonnegative Latent Factor Model

Yurong Zhong, and Xin Luo, *Senior Member*, *IEEE*

**Abstract**—An alternating-direction-method-based nonnegative latent factor model can perform efficient representation learning to a high-dimensional and incomplete (HDI) matrix. However, it introduces multiple hyper-parameters into the learning process, which should be chosen with care to enable its superior performance. Its hyper-parameter adaptation is desired for further enhancing its scalability. Targeting at this issue, this paper proposes an Adaptive Alternating-direction-method-based Nonnegative Latent Factor ($A^2$NLF) model, whose hyper-parameter adaptation is implemented following the principle of particle swarm optimization. Empirical studies on nonnegative HDI matrices generated by industrial applications indicate that $A^2$NLF outperforms several state-of-the-art models in terms of computational and storage efficiency, as well as maintains highly competitive estimation accuracy for an HDI matrix's missing data.

——————————— ◆ ———————————

## I. INTRODUCTION

LARGE SCALE INTERACTION DATA are commonly encountered in various big data-related applications, such as user preferences in recommender systems [7-9, 48] and protein interactomes in bioinformatics [13, 14]. They contain massive knowledge describing various useful patterns, e.g., potential user preferences [7-9, 48] and probable but unobserved interactomes [13, 14]. Such interactions commonly defined among numerous nodes and Observed interactions are highly sparse. Hence, they are commonly quantized into a high-dimensional and incomplete (HDI) matrix [14-16, 44]. Precise extraction of such knowledge is an essential yet challenging issue in the area of big data analysis and knowledge discovery [13-16, 31]. Great efforts have been paid to it, thus generating various models [11-21, 27-29, 31, 33, 44-46]. Among them, a nonnegative latent factor analysis (NLFA)-based approach has proven to be highly efficient [31, 33, 45, 46]. Note that different from a nonnegative matrix factorization (NMF) model, an NLFA model takes an HDI matrix as its input [31].

An NLFA model is initially proposed to solve a nonnegativity-constrained and non-convex learning objective via a single latent factor-dependent, nonnegative and multiplicative update (SLF-NMU) algorithm [31]. Note that such a learning objective is defined on an HDI matrix's known entries only, thereby enabling its computational and storage cost to be linear with an HDI matrix's density quantity measured by the ratio of known entry count to the total entry count in such a matrix. Therefore, when analyzing an HDI matrix, an NLFA model significantly outperforms traditional full-matrix-dependent models [15-21, 27-29] in terms of both computational and storage efficiencies, as well as achieves highly accurate representation to its observed data.

However, an SLF-NMU algorithm is implemented by manipulating the learning rate of additive gradient descent-based learning rules to cancel the corresponding negative terms, thereby implementing its nonnegative and multiplicative learning process [28, 31]. As discussed in [45], such an algorithm converges slowly on large-scale datasets. As a result, although its time cost per iteration is very low on an HDI matrix, it takes many training iterations to converge, thus significantly increasing its time cost to converge. Correspondingly, Luo *et al*. propose an <u>A</u>lternating-direction-method-based <u>N</u>onnegative <u>L</u>atent <u>F</u>actor (ANLF) model [33] with three ideas:

a) Additional optimization variables are introduced to split the nonnegative constraints from the least-squares fitting variables;
b) Utilizing a data density-oriented strategy on its learning objective, constraints and augmentation terms;
c) Adopting the principle of an alternating-direction-method of multipliers (ADMM) [32] to split the whole optimization task into several tasks to form an effective solving sequence, where each task is solved based on the latest state of those just solved.

With such a design, an ANLF model enjoys its fast convergence rate and high representation learning ability [33]. However, an ANLF model's performance depends heavily on its hyper-parameters, i.e., augmentation coefficient and learning rate. Commonly, they should be carefully tuned via manual grid-search [1], which leads to high computational cost. To address this issue, this paper proposes an <u>A</u>daptive <u>A</u>lternating-direction-method-based <u>N</u>onnegative Latent Factor ($A^2$NLF) model. Its main principle is to implement hyper-parameter adaptation in an ANLF model following the principle of particle swarm optimization (PSO). Empirical

———————————————————————————————
- *Corresponding author: Xin Luo.*
- *Y. Zhong and X. Luo are with the School of Computer Science and Technology, Dongguan University of Technology, Dongguan, Guangdong 523808, China (e-mail: zhongyurong91@gmail.com, luoxin21@gmail.com).*
- *X. Luo is also with the Chongqing Institute of Green and Intelligent Technology, Chinese Academy of Sciences, Chongqing 400714, China.*

studies on four industrial HDI matrices indicate that an A²NLF model achieves high efficiency in both computation and storage, as well as accurate representation to an HDI matrix. Section II gives the preliminaries, Section III presents an A²NLF model, Section IV provides the empirical studies, and finally, Section V concludes this paper.

## II. PRELIMINARIES

### A. Problem of NLFA on an HDI matrix

A nonnegative HDI matrix $Y$ is the fundamental input in our scenes as defined in [10, 14, 30, 31].

**Definition 1.** Given $U$ and $I$, $Y^{|U|\times|I|}$ quantizes certain interactions among them. Given $\Lambda$ and $\Gamma$ of $Y$, $Y$ is HDI if $|\Lambda|\ll|\Gamma|$.

Given $Y$ and $\Lambda$, $Y$'s rank-$d$ approximation $\hat{Y}=AX^{\mathrm{T}}$ is what an NLFA model seeks for as $A$ and $X\geq 0$. To acquire $A$ and $X$, an objective function is defined on $\Lambda$ only for connecting $Y$ with $\hat{Y}$. With the Euclidean distance, it is given as [6, 11, 25-28, 31, 47]:

$$\varepsilon = \frac{1}{2}\sum_{y_{u,i}\in\Lambda}\left(y_{u,i}-\sum_{k=1}^{d}a_{u,k}x_{i,k}\right)^{2}, \quad (1)$$
$$s.t.\quad A\geq 0, X\geq 0.$$

Note that (1) adopts the more concise form without considering the linear biases [6, 45] or an extended Latent Factor (LF) space [5, 33, 46]. However, an extended model shares the same principle with (1), making it compatible with the following methodology.

## III. AN A²NLF MODEL

### A. Learning Objective

To solve (1), we further introduce additional optimization variables $P^{|U|\times d}$ and $Z^{|I|\times d}$ to split the nonnegative constraints from the least-squares fitting variables, thus extending (1) to:

$$\varepsilon = \frac{1}{2}\sum_{y_{u,i}\in\Lambda}\left(y_{u,i}-\sum_{k=1}^{d}p_{u,k}z_{i,k}\right)^{2}, \quad (2)$$
$$s.t.\quad a_{u,k}=p_{u,k}, a_{u,k}\geq 0, x_{i,k}=z_{i,k}, x_{i,k}\geq 0.$$

Afterwards, an augmented Lagrangian function $g$ corresponding to (2) is built in a single-element-dependent way [33],

$$g = \frac{1}{2}\sum_{y_{u,i}\in\Lambda}\left(y_{u,i}-\sum_{k=1}^{d}p_{u,k}z_{i,k}\right)^{2} + \sum_{(u,k)}h_{u,k}\left(p_{u,k}-a_{u,k}\right) + \sum_{(i,k)}w_{i,k}\left(z_{i,k}-x_{i,k}\right)$$
$$+ \sum_{(u,k)}\frac{\lambda|\Lambda(u)|}{2}\left(p_{u,k}-a_{u,k}\right)^{2} + \sum_{(i,k)}\frac{\lambda|\Lambda(i)|}{2}\left(z_{i,k}-x_{i,k}\right)^{2}, \quad (3)$$

### B. ADMM-incorporated Learning Scheme

$\forall u\in U$, $i\in I$, $k\in\{1,\ldots,d\}$, by updating a single parameter while alternatively fixing the others, we have the following rules:

$$p_{u,k} \leftarrow \frac{\sum_{i\in\Lambda(u)}z_{i,k}\left(y_{u,i}-\sum_{j=1,j\neq k}^{d}p_{u,j}z_{i,j}\right)+\lambda|\Lambda(u)|a_{u,k}-h_{u,k}}{\sum_{i\in\Lambda(u)}(z_{i,k})^{2}+\lambda|\Lambda(u)|}, \quad (4a)$$

$$z_{i,k} \leftarrow \frac{\sum_{u\in\Lambda(i)}p_{u,k}\left(y_{u,i}-\sum_{j=1,j\neq k}^{d}p_{u,j}z_{i,j}\right)+\lambda|\Lambda(i)|x_{i,k}-w_{i,k}}{\sum_{u\in\Lambda(i)}(p_{u,k})^{2}+\lambda|\Lambda(i)|}, \quad (4b)$$

$$a_{u,k} \leftarrow \max\left(0, p_{u,k}+\frac{h_{u,k}}{\lambda|\Lambda(u)|}\right), \quad (4c)$$

$$x_{i,k} \leftarrow \max\left(0, z_{i,k}+\frac{w_{i,k}}{\lambda|\Lambda(i)|}\right), \quad (4d)$$

$$h_{u,k} \leftarrow h_{u,k}+\eta\lambda|\Lambda(u)|\left(p_{u,k}-a_{u,k}\right), \quad (4e)$$

$$w_{i,k} \leftarrow w_{i,k}+\eta\lambda|\Lambda(i)|\left(z_{i,k}-x_{i,k}\right), \quad (4f)$$

Afterwards, a training iteration of learning objective (3) is split into $d$ disjoint tasks where each task contains three subtasks. Given $k\in\{1,\ldots,d\}$, an ADMM-based learning sequence in the $k$-th task is implemented as follows,

**a) Subtask One:**

$$\begin{cases} P_{.,k}^{t+1} \overset{(4a)}{\leftarrow} \underset{P_{.,k}}{\arg\min}\, g\left(\left[P_{.,1\sim(k-1)}^{t+1}, P_{.,k}, P_{.,(k+1)\sim d}^{t}\right],\left[Z_{.,1\sim(k-1)}^{t+1}, Z_{.,k\sim d}^{t}\right], A_{.,k}^{t}, X_{.,k}^{t}, H_{.,k}^{t}, W_{.,k}^{t}\right), \\ Z_{.,k}^{t+1} \overset{(4b)}{\leftarrow} \underset{Z_{.,k}}{\arg\min}\, g\left(\left[P_{.,1\sim(k-1)}^{t+1}, P_{.,k\sim d}^{t}\right],\left[Z_{.,1\sim(k-1)}^{t+1}, Z_{.,k}, Z_{.,(k+1)\sim d}^{t}\right], A_{.,k}^{t}, X_{.,k}^{t}, H_{.,k}^{t}, W_{.,k}^{t}\right); \end{cases} \quad (5a)$$

**b) Subtask Two:**

$$\begin{cases} A_{.,k}^{t+1} \overset{(4c)}{\leftarrow} \underset{A_{.,k}\geq 0}{\arg\min}\, g(P_{.,k}^{t+1}, Z_{.,k}^{t+1}, A_{.,k}, X_{.,k}^{t}, H_{.,k}^{t}, W_{.,k}^{t}), \\ X_{.,k}^{t+1} \overset{(4d)}{\leftarrow} \underset{X_{.,k}\geq 0}{\arg\min}\, g(P_{.,k}^{t+1}, Z_{.,k}^{t+1}, A_{.,k}^{t}, X_{.,k}, H_{.,k}^{t}, W_{.,k}^{t}); \end{cases} \quad (5b)$$

**c) Subtask Three:**

$$\begin{cases} H_{.,k}^{t+1} \overset{(4e)}{\leftarrow} H_{.,k}^{t} + \eta\lambda\left|\Lambda(u)\right|\nabla_{H}g\left(P_{.,k}^{t+1}, Z_{.,k}^{t+1}, A_{.,k}^{t+1}, X_{.,k}^{t+1}, H_{.,k}^{t}, W_{.,k}^{t}\right), \\ W_{.,k}^{t+1} \overset{(4f)}{\leftarrow} W_{.,k}^{t+1} + \eta\lambda\left|\Lambda(i)\right|\nabla_{W}g\left(P_{.,k}^{t+1}, Z_{.,k}^{t+1}, A_{.,k}^{t+1}, X_{.,k}^{t+1}, H_{.,k}^{t}, W_{.,k}^{t}\right), \end{cases} \quad (5c)$$

where the actual optimization of each parameter follows the learning rules given in (4). From (5), $P$, $Z$, $A$, $X$, $H$ and $W$ are designed to be updated in a column-wise way, i.e., the $k$-th task takes care of parameters in their $k$-th column as the others are alternatively fixed [21-23, 32, 33, 37-39].

*C. Hyper-parameter Adaptation*

Following prior research [3, 4, 34, 35], we adopt the PSO principle to implement hyper-parameter adaptation. Firstly, a swarm consisting of $Q$ particles is built, where the $q$-th particle maintains a 2-dimension vector given as:

$$\begin{cases} v_{(q)} = \begin{bmatrix} v_{\lambda(q)} & v_{\eta(q)} \end{bmatrix}^{T}, \\ s_{(q)} = \begin{bmatrix} \lambda_{(q)} & \eta_{(q)} \end{bmatrix}^{T}. \end{cases} \quad (6)$$

Then, considering the update of $v_{(q)}$ and $s_{(q)}$ at the $t$-th iteration, the evolution of the $q$-th particle is implemented as:

$$\begin{aligned} v_{(q)}^{t} &= wv_{(q)}^{t-1} + b_{1}r_{1}\left(\hat{P}_{(q)}^{t-1} - s_{(q)}^{t-1}\right) + b_{2}r_{2}\left(\hat{G}^{t-1} - s_{(q)}^{t-1}\right), \\ s_{(q)}^{t} &= s_{(q)}^{t-1} + v_{(q)}^{t}, \end{aligned} \quad (7)$$

Note that $\hat{P}_{(q)}^{t-1}$ and $\hat{G}^{t-1}$ are given as:

$$\begin{cases} \hat{P}_{(q)}^{t-1} = \begin{bmatrix} \lambda_{*(q)}^{t-1} & \eta_{*(q)}^{t-1} \end{bmatrix}^{T}, \\ \hat{G}^{t-1} = \begin{bmatrix} \lambda_{*}^{t-1} & \eta_{*}^{t-1} \end{bmatrix}^{T}. \end{cases} \quad (8)$$

Note that $\lambda_{(q)}^{t}$ and $\eta_{(q)}^{t}$ denote the best $\lambda$ and $\eta$ of the $q$-th particle at the $t$-th iteration, $\lambda_{*}^{t}$ and $\eta_{*}^{t}$ denote the best $\lambda$ and $\eta$ of the whole swarm at the $t$-th iteration, respectively.

For making the whole swarm evolve to optimize the learning objective, $\hat{P}_{(q)}^{t}$ and $\hat{G}^{t}$ are updated as:

$$\hat{P}_{(q)}^{t} = \begin{cases} \hat{P}_{(q)}^{t-1}, F_{(q)}^{t} \leq F_{(q)}^{t-1} \\ s_{(q)}^{t}, F_{(q)}^{t} > F_{(q)}^{t-1} \end{cases}, \hat{G}^{t} = \begin{cases} \hat{G}^{t-1}, F_{(q)}^{t} \leq F_{(q)}^{t-1} \\ s_{(q)}^{t}, F_{(q)}^{t} > F_{(q)}^{t-1} \end{cases}, \quad (9)$$

where the fitness function $F_{(q)}^{t}$ ($t\neq 1$) is given as:

$$F_{(1)}^{t} = F_{(Q)}^{t-1}; \forall q \in \{1, ..., Q\}: F_{(q)}^{t} = \frac{M_{(q)}^{t} - M_{(q-1)}^{t}}{\hat{f}^{t} - \hat{f}^{t-1}}. \quad (10)$$

where $\hat{f}^{1}$ is initialized as a large value, e.g., 100, and updated as:

$$\hat{f}^{t} = \begin{cases} M_{(q)}^{t}, M_{(q)}^{t} < \hat{f}^{t-1}, \\ \hat{f}^{t-1}, M_{(q)}^{t} \geq \hat{f}^{t-1}. \end{cases} \quad (11)$$

Note that $M_{(q)}^t$ quantizes the generalized loss of the whole swarm. To ensure that the hyper-parameter adaptation makes A2NLF to optimize its learning objective, prediction error, $M_{(q)}^t$ is designed to be:

$$M_{(q)}^t = \sqrt{\left(\sum_{y_{u,i}\in\Omega}(y_{u,i}-\hat{y}_{u,i(q)}^t)^2\right)\Big/|\Omega|}, \text{ or } \left(\sum_{y_{u,i}\in\Omega}\left|y_{u,i}-\hat{y}_{u,i(q)}^t\right|\right)\Big/|\Omega|, \quad (12)$$

where $\Omega$ denotes the validation set disjoint with the training set $\Lambda$ and testing set K, and $\hat{y}_{u,i(q)}^t$ denotes the prediction generated by the achieved LFs with the hyper-parameter settings corresponding to the $q$-th particle at the $t$-th iteration. The $q$-th particle's position and velocity are bounded for preventing them jumping out of the searching range:

$$v_{(q)}^{t+1} = \begin{cases} \hat{v}, v_{(q)}^{t+1} > \hat{v} \\ \breve{v}, v_{(q)}^{t+1} < \breve{v} \end{cases}, s_{(q)}^{t+1} = \begin{cases} \hat{s}, s_{(q)}^{t+1} > \hat{s} \\ \breve{s}, s_{(q)}^{t+1} < \breve{s} \end{cases}, \quad (13)$$

where $\hat{v} = \begin{bmatrix} \hat{v}_\lambda & \hat{v}_\eta \end{bmatrix}^T$ and $\breve{v} = \begin{bmatrix} \breve{v}_\lambda & \breve{v}_\eta \end{bmatrix}^T$ decide the boundary of $v$, $\hat{s} = \begin{bmatrix} \hat{\lambda} & \hat{\eta} \end{bmatrix}^T$ and $\breve{s} = \begin{bmatrix} \breve{\lambda} & \breve{\eta} \end{bmatrix}^T$ decide the boundary of $\lambda$ and $\eta$, $\hat{\lambda}$ and $\breve{\lambda}$ denote upper and lower bounds for $\lambda$, and $\hat{\eta}$, $\breve{\eta}$ denote upper and lower bounds for $\eta$.

## IV. EXPERIMENTAL RESULTS AND ANALYSIS

### A. General Settings

**Datasets.** Four nonnegative HDI matrices are adopted in our experiments, whose details are given in Table I. On each dataset, the known entry set is randomly divided into ten disjoint subsets. Each time we select seven subsets as the training set $\Lambda$ to train the model, one as the validation set $\Omega$ to monitor the training process, and the remaining two as the testing set K to test the performance of a resultant model. The above process is sequentially repeated for ten times to achieve a ten-fold cross validation process. The averaged results and standard deviations are carefully recorded [6-8, 16, 24].

TABLE I. Details of Adopted Datasets.

| No. | Name | $|\Lambda|+|\Gamma|$ | $|U|$ | $|I|$ | Source |
|---|---|---|---|---|---|
| D1 | EM | 2811718 | 61265 | 1623 | EachMovie |
| D2 | Jester | 1186324 | 16384 | 100 | Jester [35] |
| D3 | ML1M | 1000209 | 6040 | 3706 | MovieLens [43] |
| D4 | ML10M | 10000054 | 71567 | 65133 | MovieLens [43] |

**Accuracy Metric.** Root mean squared error (RMSE) [6-8, 16, 40-43] are frequently adopted to measure accuracy of estimating missing data:

$$\text{RMSE} = \sqrt{\left(\sum_{y_{u,i}\in K}(y_{u,i}-\hat{y}_{u,i})^2\right)\Big/|K|}.$$

It is evident to find that lower RMSE is corresponding to higher estimation accuracy of missing data on K.

**Model Settings.** Four models are involved in the experiments, whose details are given in Table II. Moreover, the training process of a tested model terminates if 1) the difference in training error between two consecutive iterations becomes smaller than $10^{-5}$; or the training error continually increases in five consecutive iterations; and 2) the iteration count reaches a preset threshold, i.e. 1000.

TABLE II. Details of Tested Models.

| No. | Name | Description |
|---|---|---|
| M1 | ANLF | An original ANLF model [33]. |
| M2 | NMC | A nonnegative matrix completion model relying on ADMM whose optimization task is split into four tasks [21]. |
| M3 | I-AutoRec | An LF analysis model based on an item-oriented auto-encoder paradigm [49]. |
| M4 | A²NLF | The adaptive model proposed in this study. |

### B. Comparison against State-of-the-art Models

The comparison results are summarized in Tables III and IV. The RMSE of M1-4 on D1-4 are respectively given in Table III. The total time costs of M1-4 on D1-4 are recorded in Table IV. From them, we have the following important findings:

a) **M4's hyper-parameter adaptation does not impair its prediction accuracy for missing data of an HDI matrix.** In comparison with the original ANLF model, i.e., M1, M4 achieves lower RMSE on D1-2 and D4, as shown in Table III. From this point of view, the hyper-parameter adaptation mechanism of M4 is highly efficient to enable its fine convergence on an HDI matrix. On all four testing cases, M4's prediction accuracy ranks the first on two cases (i.e., RMSE on D2 and D4), and the second on the remaining two case (i.e., RMSE on D1 and D3). Hence, M4's prediction accuracy is highly competitive with its hyper-parameter adaptation mechanism.

b) **M7's computational efficiency is significantly higher than that of its peers.** For example, as shown in Table IV, the total time cost of M4 on D3 is 26 seconds, which is about 96.57%, 99.96%, 99.94%, 96.37%, 99.69% and 99.74% less than M1's 759, M2's 71927, M3's 45849, M4's 717, M5's 8359 and M6's 9921. The same results are also observed on the remaining three testing cases. The main reason for these positive results should be owing to M7's hyper-parameter adaptation.

TABLE III. RMSE of M1-4 on D1-4.

|    | D1 | D2 | D3 | D4 |
|----|----|----|----|----|
| M1 | 0.2373±2.2E-6/4 | 1.0187±1.1E-6/2 | **0.8665±7.8E-4/1** | 0.8096±2.9E-6/2 |
| M2 | 0.2384±1.0E-4/5 | 1.0787±8.6E-7/4 | 0.8713±3.7E-4/3 | Failure |
| M3 | **0.2302±2.6E-3/1** | 1.1494±4.1E-5/6 | 0.8845±1.1E-3/4 | 0.8436±1.2E-3/3 |
| M4 | 0.2339±2.7E-4/2 | **1.0172±7.4E-4/1** | 0.8675±8.2E-4/2 | **0.8089±9.4E-4/1** |

TABLE IV. Total Time Cost of M1-4 in RMSE on D1-4 (Seconds).

|    | D1 | D2 | D3 | D4 |
|----|----|----|----|----|
| M1 | 434±25/2 | 275±47/5 | 759±87/3 | 2,972±378/2 |
| M2 | 209,670±36,625/7 | 138±39/4 | 45,849±2,479/6 | Failure[1] |
| M3 | 10,804±832/4 | 2,109±295/7 | 8,359±532/4 | 266,885±12,775/3 |
| M4 | **46±4/1** | **25±5/1** | **26±6/1** | **334±46/1** |

## V. CONCLUSIONS

This paper proposes an $A^2NLF$ model for efficient representation learning to an HDI matrix generated by various industrial applications. Its hyper-parameter adaptation mechanism is implemented by following the PSO principle, which ensures its representation learning ability as well as high computational efficiency.

Next, we aim to implement more effective hyper-parameter adaptation strategies. They can be achieved by adopting advanced evolutionary computation algorithm for improving the representation ability to an HDI matrix with high efficiency in both computation and storage [2, 36]. It is also desired to figure out the solution type achieved by an $A^2NLF$ model on an HDI matrix.